\documentclass[letterpaper, 10 pt, conference]{ieeeconf}  % Comment this line out if you need a4paper

\IEEEoverridecommandlockouts                              % This command is only needed if 
\usepackage{graphicx}
\usepackage{epsfig} % for postscript graphics files
\usepackage{amsmath} % assumes amsmath package installed
\usepackage{amssymb}  % assumes amsmath package installed

\usepackage{xcolor}
\definecolor{linkcolor}{RGB}{6,69,173}
\usepackage[
colorlinks=true,allcolors=linkcolor,pageanchor=true,plainpages=false,pdfpagelabels,bookmarks,bookmarksnumbered,backref=page
]{hyperref}

\usepackage{float}
\usepackage{capt-of}
\usepackage{balance}
\usepackage[nameinlink,capitalise]{cleveref}
\crefname{figure}{Fig.}{Figs.}

\newcommand{\x}{\mathbf{x}}
\newcommand{\z}{\mathbf{z}}
\newcommand{\pc}{\mathbf{P}}
\newcommand{\G}{\mathbf{G}}

\usepackage{xspace}
\newcommand{\ncf}{\texttt{NCF}\xspace}

\let\oldtwocolumn\twocolumn
\renewcommand\twocolumn[1][]{%
    \oldtwocolumn[{#1}{
    \begin{center}
        \vspace{-6mm}
        \includegraphics[scale=0.24]{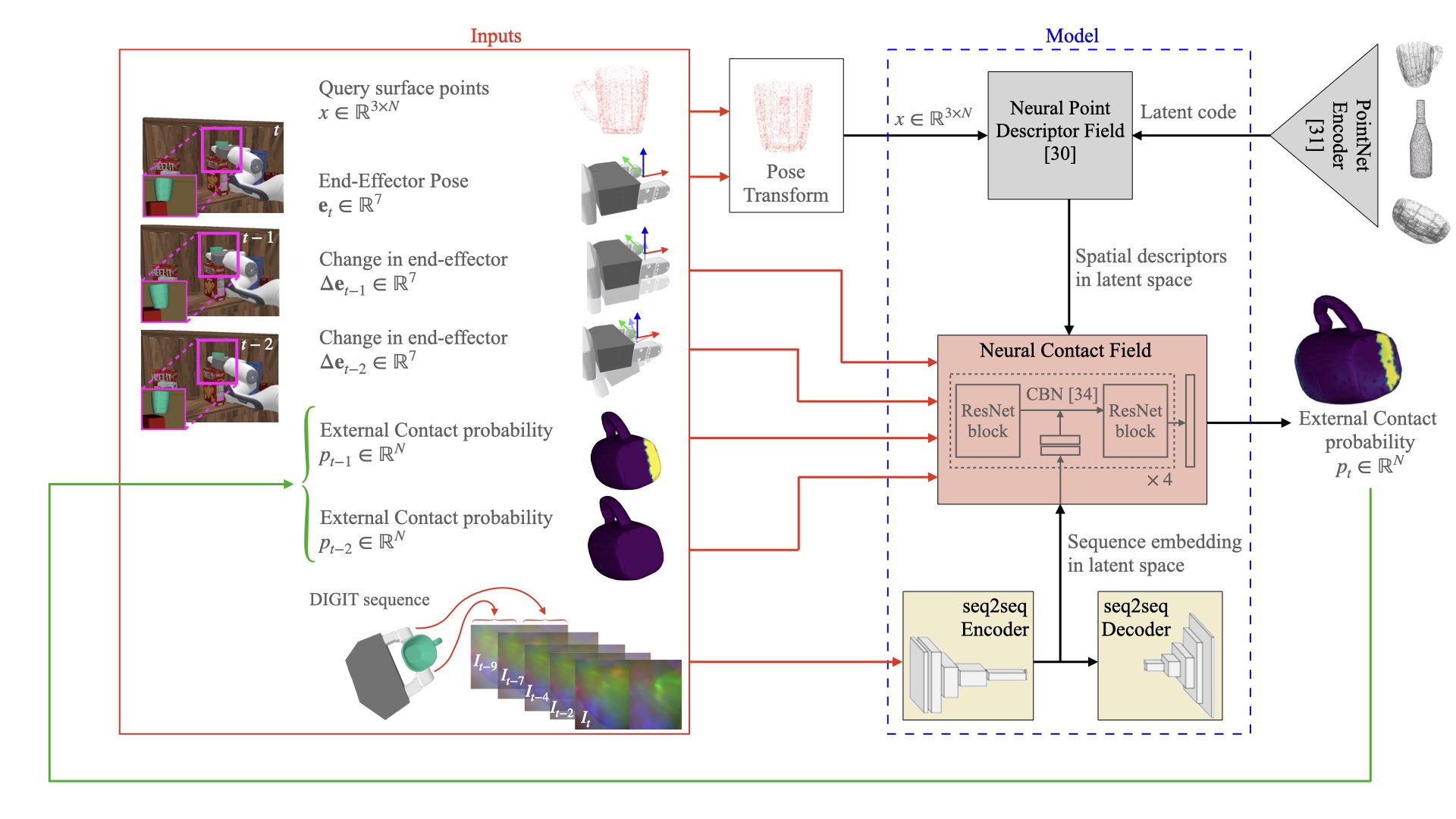}
        \vspace{-6mm}
        \captionof{figure}{Neural Contact Fields (\ncf) is an implicit representation for tracking extrinsic contact on an object surface (between object and environment) with vision-based tactile sensing (between robot hand and object). Our model outputs extrinsic contact probability at the current timestep for a set of query points on the 3D surface of the object. This is done by taking a history of end-effector poses, tactile images and extrinsic contact probabilities~as~inputs.
        }
       \label{fig:neural_contact_field_arq}
    \end{center}
    }]
}

\title{\LARGE \bf
Neural Contact Fields: Tracking Extrinsic Contact with Tactile Sensing
}

\author{Carolina Higuera$^{1}$, Siyuan Dong$^{1}$, Byron Boots$^{1}$, and Mustafa Mukadam$^{2}$\\[2mm]
$^{1}$University of Washington, $^{2}$Meta AI
}

\begin{document}

\maketitle

% \begin{figure*}[!h]
%     \includegraphics[scale=0.26]{figures/neural_contact_fields.jpeg}
%     \captionof{figure}{Neural Contact Fields is an implicit representation for tracking extrinsic contact on an object surface based on vision-based tactile sensing. Our framework takes in a set of query points on the 3D surface of the object, a history of tactile images, end-effector poses, and contact patches, and outputs the probability of extrinsic contact for each query point.  We leverage on neural implicit representations and neural descriptor fields to extract category-level spatial descriptors in latent space that encode the relationship of the 3D point with salient features of the object.}
%     \label{fig:neural_contact_field_arq}
% \end{figure*}

\thispagestyle{empty}
\pagestyle{empty}

%%%%%%%%%%%%%%%%%%%%%%%%%%%%%%%%%%%%%%%%%%%%%%%%%%%%%%%%%%%%%%%%%%%%%%%%%%%%%%%%
\begin{abstract}
We present Neural Contact Fields, a method that brings together neural fields and tactile sensing to address the problem of tracking extrinsic contact between object and environment. Knowing where the external contact occurs is a first step towards methods that can actively control it in facilitating downstream manipulation tasks. Prior work for localizing environmental contacts typically assume a contact type (e.g. point or line), does not capture contact/no-contact transitions, and only works with basic geometric-shaped objects. Neural Contact Fields are the first method that can track arbitrary multi-modal extrinsic contacts without making any assumptions about the contact type. Our key insight is to estimate the probability of contact for any 3D point in the latent space of object’s shapes, given vision-based tactile inputs that sense the local motion resulting from the external contact. In experiments, we find that Neural Contact Fields are able to localize multiple contact patches without making any assumptions about the geometry of the contact, and capture contact/no-contact transitions for known categories of objects with unseen shapes in unseen environment configurations. In addition to Neural Contact Fields, we also release our YCB-Extrinsic-Contact dataset of simulated extrinsic contact interactions to enable further research in this area. Project page: \url{https://github.com/carolinahiguera/NCF}
\end{abstract}

%%%%%%%%%%%%%%%%%%%%%%%%%%%%%%%%%%%%%%%%%%%%%%%%%%%%%%%%%%%%%%%%%%%%%%%%%%%%%%%%
\section{Introduction}

We investigate the problem of tracking extrinsic contact between object and environment using tactile perception between robot hand and object. Consider the task of placing a book on a bookcase when the fingers are firmly grasping the book. While the book is in free space, no shear forces are experienced in the fingers. However, once the book makes external contact, for example if a corner touches the shelf, the book rotates in accordance with the kinematic and frictional constraints of the contact. As a result, shear forces are perceived in the fingers. In such manipulation tasks tracking extrinsic contact with tactile sensing becomes critical for spatial understating, since vision is heavily occluded. It is also a first step towards building methods that can then leverage as well as actively control extrinsic contacts in downstream policies.

%Inferring the location of external contact is critical for manipulation in clutter and may allow for control of the contact during downstream tasks.

Despite the valuable information that an extrinsic contact tracker can provide about object-environment interaction, for example for extrinsic dexterity~\cite{zhou2022learning, 6907062}, this problem is understudied in literature. Prior approaches for localizing environmental contacts are characterized by making hypothesis about the geometry of the contact. For example, Ma et al.~\cite{prior_work_csp} formulate the problem as constrained optimization, where different constraints must be defined to solve for parameters that localize external contact, such as position of a point, direction of a line, or normal direction of a plane. Kim et al.~\cite{prior_work_cline} uses factor graphs and tactile measurements to actively estimate the contact line between the object and its environment for a peg-in-hole insertion task, while having a controller that enforces the line contact. Although these works have started to explore the problem of extrinsic contact localization, several challenges remain. For instance, it is not clear: (i) how to track multi-modal contact patches without assuming a contact type; (ii) how to track contact to no-contact interactions, and vice versa; and (iii) how to generalize to complex object shapes and new environments.

% For instance, it is not clear how to generalize to complex object shapes and new environments, how to track multi-modal contact patches without assuming a contact type, or how to track contact to no-contact interactions, and vice versa.

In this work, we address these challenges with Neural Contact Fields (\ncf), the first method that can track arbitrary multi-modal extrinsic contact from tactile perception. Our key insight is to leverage neural fields to generalize across different object shapes, given the local motion produced by the contact, which are well-captured by vision-based tactile sensors~\cite{digit_paper}. The full architecture of Neural Contact Fields is illustrated in~\cref{fig:neural_contact_field_arq}. Our method estimates the probability of external contact for any 3D point on an object surface given a sequence of tactile images and the most recent history of end-effector poses and external contact probabilities. We train \ncf to track extrinsic contact on three categories of objects with simulated tactile data. In experiments, we find that \ncf is able to localize multiple contact patches without making any assumptions about its geometry and can capture contact/no-contact transitions, on unseen shapes in unseen environments.

In addition to open-sourcing Neural Contact Fields, we also release the YCB-Extrinsic-Contact dataset of simulated extrinsic contact interactions for three categories of objects in different environments. The dataset compiles information about end-effector poses, DIGIT tactile sensor images~\cite{digit_paper} from the TACTO simulator~\cite{tacto_sim}, and extrinsic contact ground-truth per time step of each training/testing trajectory. This dataset provides a simulation benchmark that we hope will enable further research in extrinsic contact perception. %With this, we want to provide to the robotics community a simulation benchmark and a strong extrinsic-contact baseline to encourage further research in this area.

%%%%%%%%%%%%%%%%%%%%%%%%%%%%%%%%%%%%%%%%%%%%%%%%%%%%%%%%%%%%%%%%%%%%%%%%%%%%%%%
\section{Related Work}

\subsection{External contact localization}
Although localizing external contacts can provide rich information to respond correctly in downstream tasks, such as insertion, packing, or assembling, this important problem has received little attention. Ma et al.~\cite{prior_work_csp} discuss a theoretical basis for localizing environmental contacts. They formulate the problem as a constraint-based estimation subject to the kinematic constraints imposed by tactile measurements of the object local motion, as well as the kinematic and frictional constraints imposed by rigid-body mechanics. This work presents a method to estimate three possible hypothesis of contact: point, line and patch contact. For each mode, different constraints must be formulated to solve for the parameters that allow to localize the external contact on the object, such as position of a point, direction of a line, or normal direction of a plane. To solve the problem, some assumptions must be guaranteed such as the object remains in contact with the environment and that the grasp is stable. The work acknowledges that a key limitation lies in having to assume a particular contact type to formulate the constraints and that a stronger implementation should consider multiple hypothesis for contact formations. In addition,~\cite{prior_work_cline} follows a similar approach, using factor graphs and tactile measurements to actively estimate the contact line between the object and its environment for a peg-in-hole insertion task. This method allows to parameterize the contact line, assuming that the object is always making contact, and generalizes well for objects with basic shapes and flat surfaces.

A learning based approach is presented in~\cite{prior_work_point_cloud}, in which a collision model is learned from point cloud of the object and its environment. The model predicts the likelihood that the object collides with the scene but does not provide further information about where is the contact located on the object. This method generalizes across four objects categories (mugs, cylinders, boxes, and bowls) with different scenes. 

Our Neural Contact Fields generalize across three object categories, can localize multi-modal external contacts without making any assumptions about the contact type, and can capture contact/no-contact transitions.

\subsection{Vision-based tactile sensing in robotics}
Vision-based tactile sensing has been actively used in manipulation tasks given their ability to capture local contact events with high accuracy. Generally, these sensors are located at the gripper fingers, capturing the small forces and object displacements  that allow to detect and predict contact events. For example,  \cite{tactile_contact_events}~shows that tactile sensing is adequate to distinguish slippage, rolling, making/breaking contact, among others, from a data-driven approach. Regarding their usage in manipulation tasks, \cite{tactile_policy_1} uses tactile sensing for packing four basic objects shapes in a box, with the hypothesis that different error directions will result in distinguishable tactile imprints. In~\cite{tactile_policy_2} tactile sensing is used as the state for learning an insertion policy with reinforcement learning. This work highlights the need of feedback mechanisms to interactively correct the misalignment between object and environment and is a potential application for Neural Contact Fields.

% Vision-based tactile sensing has been actively used in manipulation tasks given their ability to capture local contact events with high accuracy. Generally, these sensors are located at the gripper fingers, allowing to capture the small forces and object displacements that occur when the grasped object is making contact with the environment. For a reliable robotic grasping system and its further success in an insertion task, the system must be able to detect and predict contact events in order to respond correctly. \cite{tactile_contact_events}~shows that tactile sensing is adequate to distinguish slippage, rolling, making/breaking contact, among others, from a data-driven approach. \cite{tactile_policy_1} uses tactile sensing for packing four basic objects shapes in a box, with the hypothesis that different error directions will result in distinguishable tactile imprints. In~\cite{tactile_policy_2} tactile sensing is used as the state for learning an insertion policy with reinforcement learning. This work highlights the need of feedback mechanisms to interactively correct the miss-alignment between object and environment and is a potential application for Neural Contact Fields.

Vision-based tactile sensing has been used in many other applications for robot manipulation. For example,  estimating object poses from touch measurements~\cite{sodhi2021tactile, sodhi2021patchgraph, Suresh21tactile} and  learning a mapping of contact shapes to object poses~\cite{tactile_pose_2, tactile_pose_3}. A global localization on an object surface is presented in~\cite{suresh2022midastouch} from a vision-based touch sensor sliding. Additionally, 3D shape reconstruction from touch has been presented in~\cite{tactile_3d_recons, 8593430} and~\cite{NEURIPS2020_a3842ed7}.

\subsection{Neural representations of 3D geometries}
Our method for extrinsic contact tracking leverages neural fields to encode complex 3D geometries. Neural fields enable representing a shape's surface by a continuous volumetric field. For instance, DeepSDF~\cite{deep_sdf} learns a Signed Distance Function (SDF) across a class of shapes, implicitly encoding a shape’s boundary as the zero-level-set of the function. Occupancy Networks~\cite{Occupancy_Networks} reason about the occupancy probability of a 3D point, conditioned on the input observation of the shape (e.g., image, point cloud, etc.). Other such examples include volume density~\cite{implicit_shape_recon1, implicit_shape_recon2} and neural radiance fields~\cite{radiance_fields1, radiance_fields2, radiance_fields3}. The most significant advantage of neural fields is that they enable representing objects and scenes with infinite resolution in an efficient parameterization. They allow operating on a latent space that encodes information about the object class, as well as its most salient geometric features. We specifically take advantage of this feature to generalize across a diverse of categories of objects and shapes. 

With such advantages, neural fields are increasingly being explored and built on for robotics applications like learning policies~\cite{neural_robotics_policy}, robot self-models for space occupancy queries~\cite{neural_robotics_model}, room scale online signed distance fields for navigation~\cite{ortizisdf}, and learning transformations from demonstrations of a pick-and-place task~\cite{ndf}.

%%%%%%%%%%%%%%%%%%%%%%%%%%%%%%%%%%%%%%%%%%%%%%%%%%%%%%%%%%%%%%%%%%%%%%%%%%%%%%%
\section{Tracking Extrinsic Contact}

In this paper we focus on the problem of tracking extrinsic contact between object and environment. When the object is making external contact, it moves in accordance with the kinematic constraints of the contact. This motion can be well-captured by a sequence of images from vision-based tactile sensors. Assuming that the object is rigidly grasped, our goals are threefold: i) to track multiple contact patches without assuming the contact type, ii) capturing contact to no-contact and vice-versa transitions, and iii) to generalize contact tracking to complex and unseen object shapes and environments. To achieve these, we estimate the probability of contact for any 3D point on an object surface with the model illustrated in~\cref{fig:neural_contact_field_arq}. This model consists of three modules:
\begin{itemize}
    \item A \textit{PointNet Encoder + Neural Point Descriptor Fields} that allow us to generalize across different object shapes. This generates a spatial descriptor in the latent space of shapes for every 3D point $\x$ on an object.
    \item A \textit{Sequence-to-Sequence Autoencoder} to extract the embedding from a sequence of tactile images. This embedding encodes the motion of the object due to the contact interaction with the environment.
    \item \textit{Neural Contact Fields} (ours), which can estimate the probability of contact for a spatial descriptor, given the embedding of the motion due to the external contact and the most recent history of end-effector poses and external contact probabilities.
\end{itemize}

\subsection{Neural Point Descriptor Fields}
In this subsection we briefly provide background on Neural Point Descriptor Fields by Simeonov et al.~\cite{ndf} and how we are using them.

Neural Point Descriptor Fields represent an object as a function that maps a 3D coordinate $\x$ to a spatial descriptor, based on the architecture of Occupancy Networks~\cite{Occupancy_Networks}. The motivation behind Neural Point Descriptor Fields is that occupancy networks can be viewed as a classifier $\Phi$, where for 3D shape reconstruction the decision boundary implicitly represents the object’s surface. By conditioning the model on different low-dimensional latent codes, occupancy networks can reconstruct different shapes. These latent codes can be obtained as the output of a PointNet encoder $\mathcal{E}$~\cite{pointnet} that takes as input the point cloud $\pc$ of the shape. The final layer performs a coarse classification for $\x$, inside or outside the shape, whereas the inner layers encode an increasing degree of detail about the surface shape.

% In occupancy networks, a point $\x$ is marked as occupied if $\Phi(\x, \mathcal{E}(\pc))$ is bigger or equal to some threshold $\tau$, and this threshold determines the thickness of the extracted 3D surface.  Therefore, the final layer performs a coarse classification for $\x$, inside or outside the shape, whereas the inner layers encode an increasing degree of detail about the surface shape.

Based on these insights, it uses the concatenation of all  activations across layers of the occupancy network as the spatial descriptor $\z$ for a 3D point $\x$. Furthermore, given that the model is conditioned on the latent code $\mathcal{E}(\pc)$, the model is forced to parameterize the spatial descriptors grounded in the latent code of the object's category:

\begin{equation}
    \label{eq:func_ndf}
    \z = g(\x|\pc) = \bigoplus_{i=1}^{L} \Phi^{i}(\x, \mathcal{E}(\pc)),
\end{equation}
where $\Phi^{i}$ is the output after activation of the $i$-th layer  with $L$ total number of layers, and $\bigoplus$ as the concatenation operator.

For tracking extrinsic contact, for each $\x$ in the set of $N$ query surface points we have spatial descriptors that encode their relationship  with salient features of the object. However, given that the object is in motion (for example, a robot arm is placing the object on a shelf) it is important that these descriptors remain unchanged regardless of the pose of the object. 

% In our work, we assume that the object is rigidly grasped, so that the configuration of the object in world frame is subject to a rigid body transform $(\mathbf{R}, \mathbf{t}) \in SE(3)$, which in turn is determined by the end-effector pose. Neural point descriptor fields allow rotation equivariance by using an occupancy network equipped with Vector Neurons~\cite{vector_neurons}, which extends classical scalar neurons to 3D vectors and guarantees SO(3) equivariance. Translation equivariance is easily implemented by mean-centering the object's point cloud $\pc$. In this way, our Neural Contact Fields works in the latent space of spatial descriptors $\z \in \mathbb{R}^n$: 

In our work, we assume that the object is rigidly grasped, so that the configuration of the object in world frame is subject to a rigid body transform $(\mathbf{R}, \mathbf{t}) \in SE(3)$, which in turn is determined by the end-effector pose. Neural point descriptor fields allow rotation equivariance by using an occupancy network equipped with Vector Neurons~\cite{vector_neurons}. Translation equivariance is easily implemented by mean-centering the object's point cloud $\pc$. In this way, our Neural Contact Fields works in the latent space of spatial descriptors $\z \in \mathbb{R}^n$: 
\begin{equation}
    \label{eq:pose_descriptors}
    \z_i = g(\x_i|\pc) = g(\mathbf{R}\x_i+\mathbf{t}|\mathbf{R}\pc+\mathbf{t}) \forall \; i \in \{ 0, N\}
\end{equation}

\subsection{Sequence-to-Sequence Autoencoder}
In our setting, the object is grasped by a robot with a parallel gripper and tactile sensors in the fingers. If the grasp is rigid and no contact is happening, the pose of the object is completely determined by the end-effector pose and no changes are observed in the tactile measurements. However, under external contact, the object might be pivoting or slipping slightly. This motion is well captured by vision-based tactile sensors. 

We capture the information of the relative motion from a sequence $\G$  with $k$ tactile images with a sequence-to-sequence autoencoder. First, we learn an autoencoder that allow us to capture a low-dimensional representation of raw $320 \times 240$ RGB tactile images per finger.  Then, our sequence-to-sequence autoencoder uses as tokens these low-dimensional representation from left and right fingers concatenated with the end-effector pose. Both encoder and decoder have a convolutional LSTM as recurrent network~\cite{convLSTM}.  This setting allow us to learn $\mathcal{E}(\G) \in \mathbb{R}^m$, an implicit representation of tactile sequences in a self-supervised manner. This implicit representation corresponds to the last hidden state of the encoder.

\subsection{Neural Contact Fields}
Our overall pipeline, shown in~\cref{fig:neural_contact_field_arq} aims to represent the extrinsic contact of an object as a function that maps a 3D point $\x$ to a probability of contact, given the object's motion captured by a sequence $\G$ of tactile images from the gripper's fingers and a canonical point cloud description of the object $\pc$:
\begin{equation}
    \label{eq:occ_function}
    f(\x, \G, \pc): \mathbb{R}^3  \times \mathbb{R}^{k \times 2 \times 320 \times 240 \times 3} \times \mathbb{R}^{3 \times N} \rightarrow [0,1]
\end{equation}

Neural Contact Fields (\ncf) makes it possible to track extrinsic contacts by estimating in latent space of spatial descriptors whether a 3D point $\x$ is making extrinsic contact or not. This is conditioned on the implicit representation of the motion of the object due to contact, which is captured by tactile images at timesteps ${t, t-2, t-4, t-7, t-9}$:
\begin{equation}
    \label{eq:mapping_external_contact}
    f(\z, \mathcal{E}(\G)): \mathbb{R}^n \times \mathbb{R}^m \rightarrow [0,1]
\end{equation}

In addition, we also provide the \ncf model with temporal information about the history of the object state, such as the probability for the queried points of being in contact $p_{t-1, t-2}$ and the change in end-effector pose with respect to the current one $\Delta \mathbf{e}_{t-1, t-2}$ for the last two timesteps.

Given the similarities between Occupancy Networks and \ncf in the sense that for both models the output is a probability, we follow a similar architecture. We feed our set of inputs through four fully-connected ResNet blocks with Conditional Batch-Normalization~\cite{CBN} to condition the network on the embedding of the object's motion sequence $\mathcal{E}(\G)$. Finally, we use a fully-connected layer and apply the sigmoid as activation function to obtain contact probabilities for each 3D coordinate $\x$.

%%%%%%%%%%%%%%%%%%%%%%%%%%%%%%%%%%%%%%%%%%%%%%%%%%%%%%%%%%%%%%%%%%%%%%%%%%%%%%%
\section{The YCB-Extrinsic-Contact Dataset}\label{sec:dataset}

\begin{figure}[!t]
    \centering
    \includegraphics[scale=0.25]{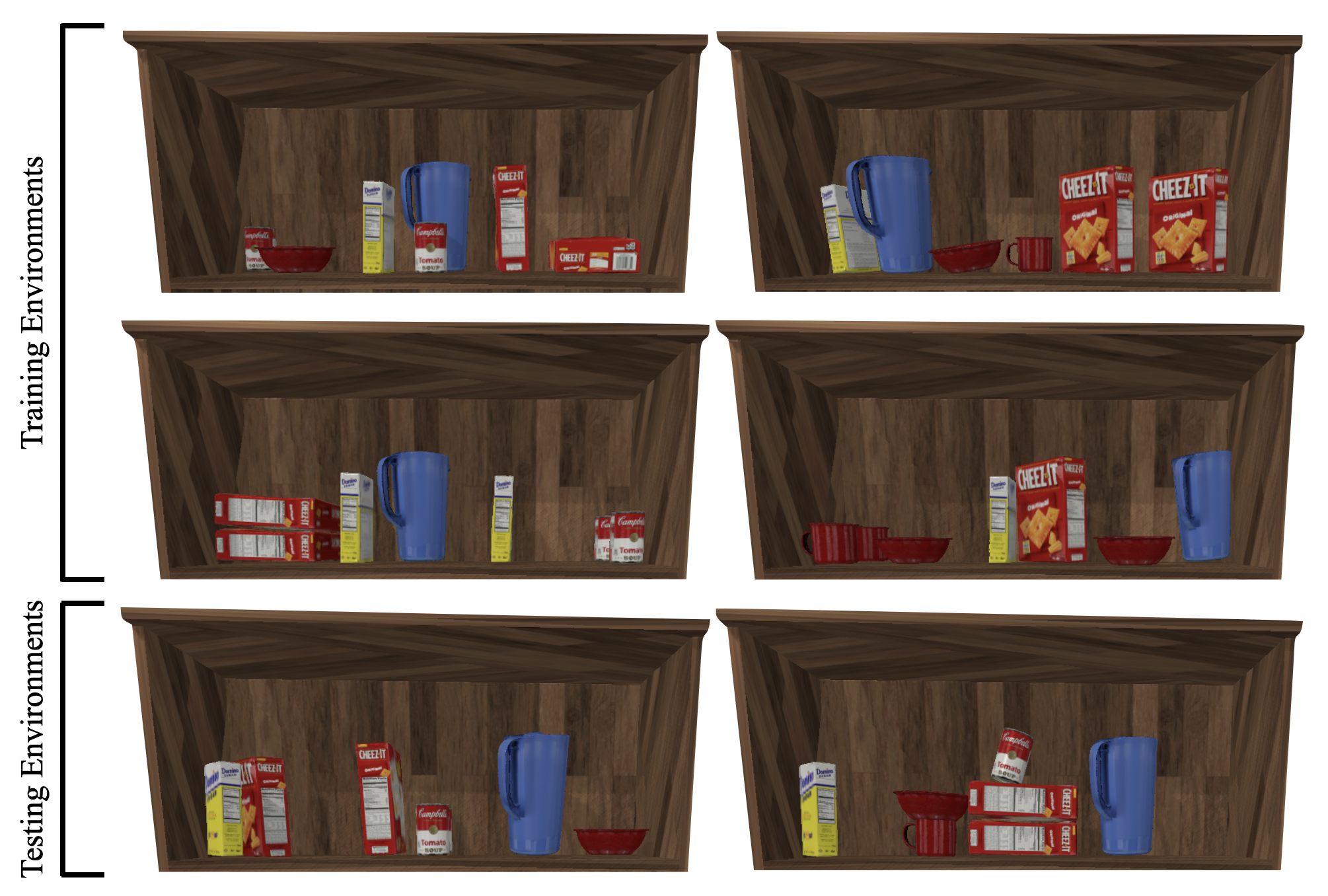}
    \vspace{-7mm}
    \caption{For collecting our YCB-Extrinsic-Contact dataset we simulate six kitchen cabinet environments with different configurations of YCB objects in PyBullet.  Four scenes are use for collecting training data and the other two for testing. Objects in these environments can move around as the robot interacts with the environment when collecting extrinsic contact data.}
    \label{fig:scenes}
    \vspace{-3mm}
\end{figure}

\begin{figure}[!t]
    \centering
    \includegraphics[scale=0.23]{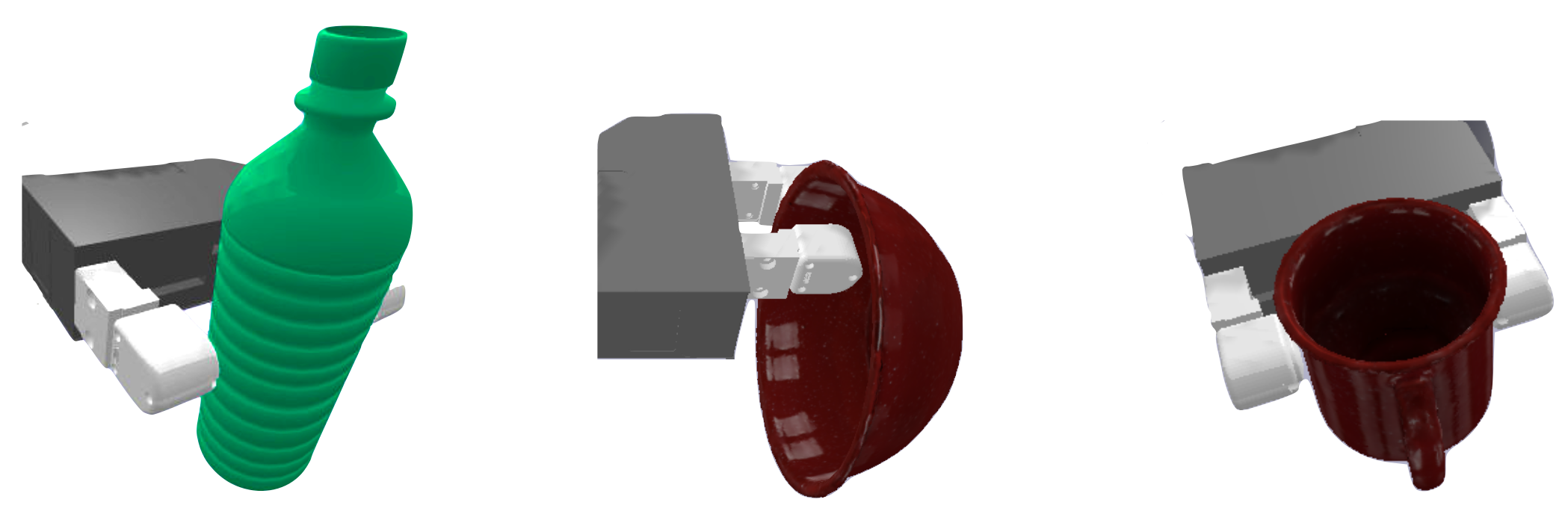}
    \vspace{-4mm}
    \caption{For collecting our dataset, the grasp pose for all bottles (left), bowls (middle), and mugs (right) is fixed and rigid as shown.}
    \label{fig:init_grasp}
    \vspace{-6mm}
\end{figure}

To evaluate our \ncf and to encourage further research in extrinsic contact modeling through tactile sensing, we open source the YCB-Extrinsic-Contact dataset. It consist of six kitchen cabinet scenarios with  YCB objects in it on different configurations, as presented in ~\cref{fig:scenes}, four for collecting training data and two for testing the model. A Franka Panda arm with a parallel gripper and DIGIT sensors~\cite{digit_paper} mounted on both fingers is interacting with these simulated environments in PyBullet~\cite{pybullet}. The trajectories that the robot follows are randomly generated. We use TACTO~\cite{tacto_sim} for simulating the sensors.  For collecting extrinsic contact data, we use three categories of objects: mugs, bottles and bowls.  We use several shapes for collecting  training trajectories, obtained from the 3D Warehouse dataset~\cite{shapes_dataset}.  In order to evaluate the generalization of our model on unseen shapes for the grasped object, we collect the testing trajectories using new shapes for the three objects in testing scenarios. The initial grasp pose for each object is fixed and rigid and a example is shown in~\cref{fig:init_grasp}. 

% mug and bowl from YCB dataset. Bottle from https://3dwarehouse.sketchup.com/model/957f897018c8c387b79156a61ad4c01/bouteille-d-eau-en-plastique

The dataset provides for every contact event, access to a history of DIGIT images for each finger, end-effector pose, a reference point cloud of the grasped object, the set of query surface points and their ground truth probabilities of external contact. 

%%%%%%%%%%%%%%%%%%%%%%%%%%%%%%%%%%%%%%%%%%%%%%%%%%%%%%%%%%%%%%%%%%%%%%%%%%%%%%%
\section{Evaluation}

\begin{figure*}[t]
    \centering
    \includegraphics[scale=0.26]{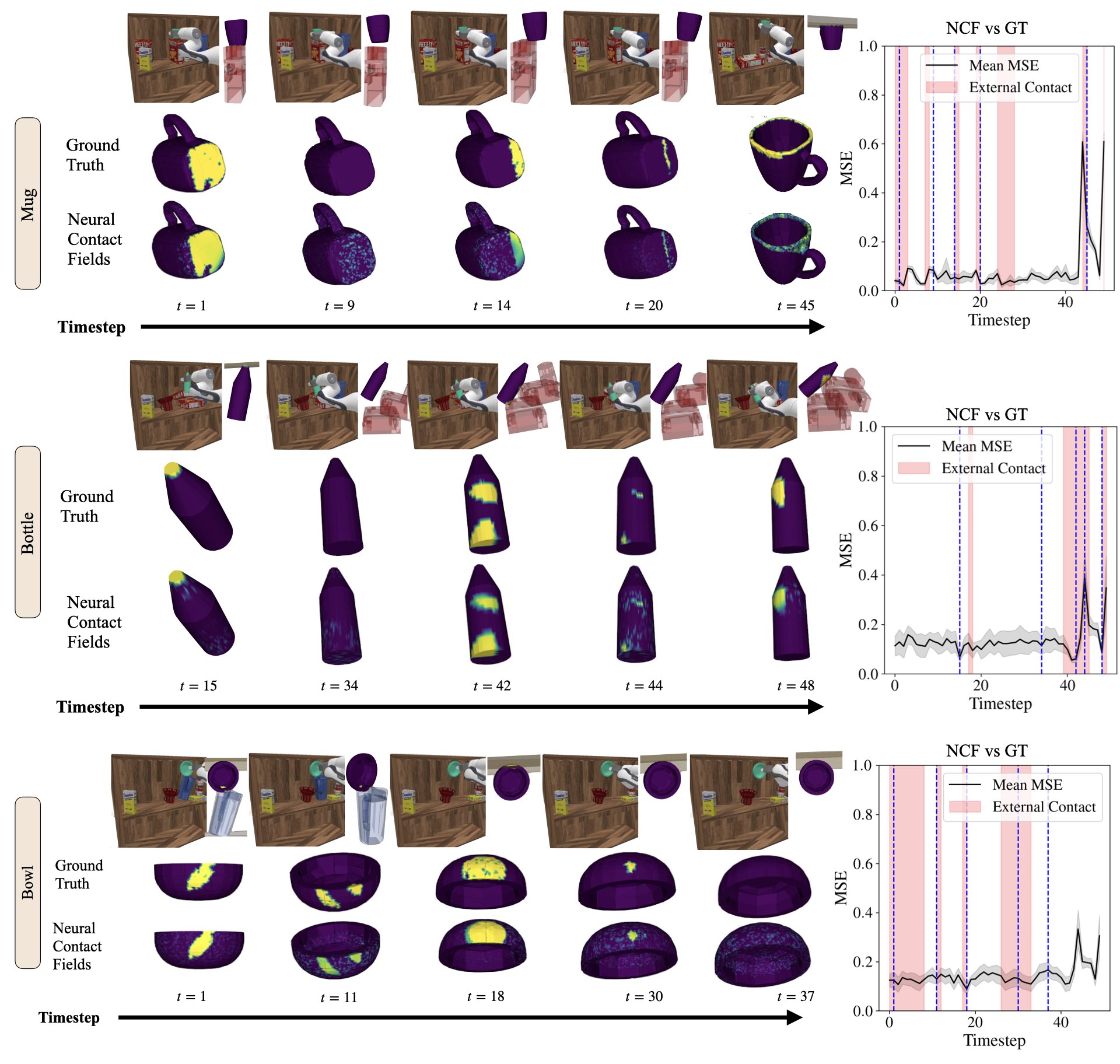}
    \vspace{-5mm}
    \caption{Snapshots of extrinsic contact predictions on simulated testing trajectories at key timesteps (blue dotted lines) for mugs, bottles and bowls. For each row: \textbf{[top]} Pybullet frame and zoom-in view of the contact interaction, \textbf{[middle]} extrinsic contact ground truth probabilities, \textbf{[bottom]} Neural Contact Fields prediction, and \textbf{[right]} MSE between the ground truth and estimated extrinsic contact probabilities for the trajectory over time.}
    \label{fig:ncf_results}
    \vspace{-6mm}
\end{figure*}

We use a pretrained implementation of Neural Point Descriptor Fields to get the spatial descriptors for 3D coordinates in the point cloud for mugs, bottles and bowls. We trained separately the sequence-to-sequence autoencoder to learn the embedding of DIGIT image sequences. These sequences have length  $k=5$ and contain the images for both fingers at timesteps $t, t-2, t-4, t-7$ and $t-9$. For our \ncf we trained four fully-connected ResNet blocks with Conditional Batch-Normalization on training trajectories that contain in total 4500 contact events. We use negative log-likelihood as loss function and Adam optimizer~\cite{adam} with learning rate $1e^{-4}$.

We evaluate \ncf offline with a GPU Nvidia RTX 3080, for which the forward pass takes $\sim$66 ms. In~\cref{fig:ncf_results} we show qualitative results of our pipeline tracking extrinsic contact. We show key snapshots of a test trajectory collected with unseen mugs, bottles, bowls, and new scenarios in simulation. From a qualitative point of view, we demonstrate the ability of \ncf to track extrinsic contacts for the three novel objects. For example, with the mug trajectory, we show that the \ncf  can transition from patch to break contact to a new contact location. With the snapshots for the trajectories with the bottle and bowl, we show that \ncf can localize multiple contact patches produced during complex contact interactions. A video of these trajectories along with the extrinsic contact tracking made by our \ncf is available in the supplementary material.

For a quantitative evaluation, we analyze the mean squared error (MSE) between the ground truth external contact probability and the predictions. At each timestep we plot the MSE after running the model 10 times and applying Monte Carlo Dropout to compute a $95\%$ confidence interval. \ncf is close to the ground truth during the majority of the trajectories. From these results, we identify cases that might induce a spike in error. For example,  when the object transitions from no-contact to contact or when the shape of the contact patch changes drastically due to a non-smooth change in end-effector pose.

\begin{figure}
    \centering
    \includegraphics[scale=0.2]{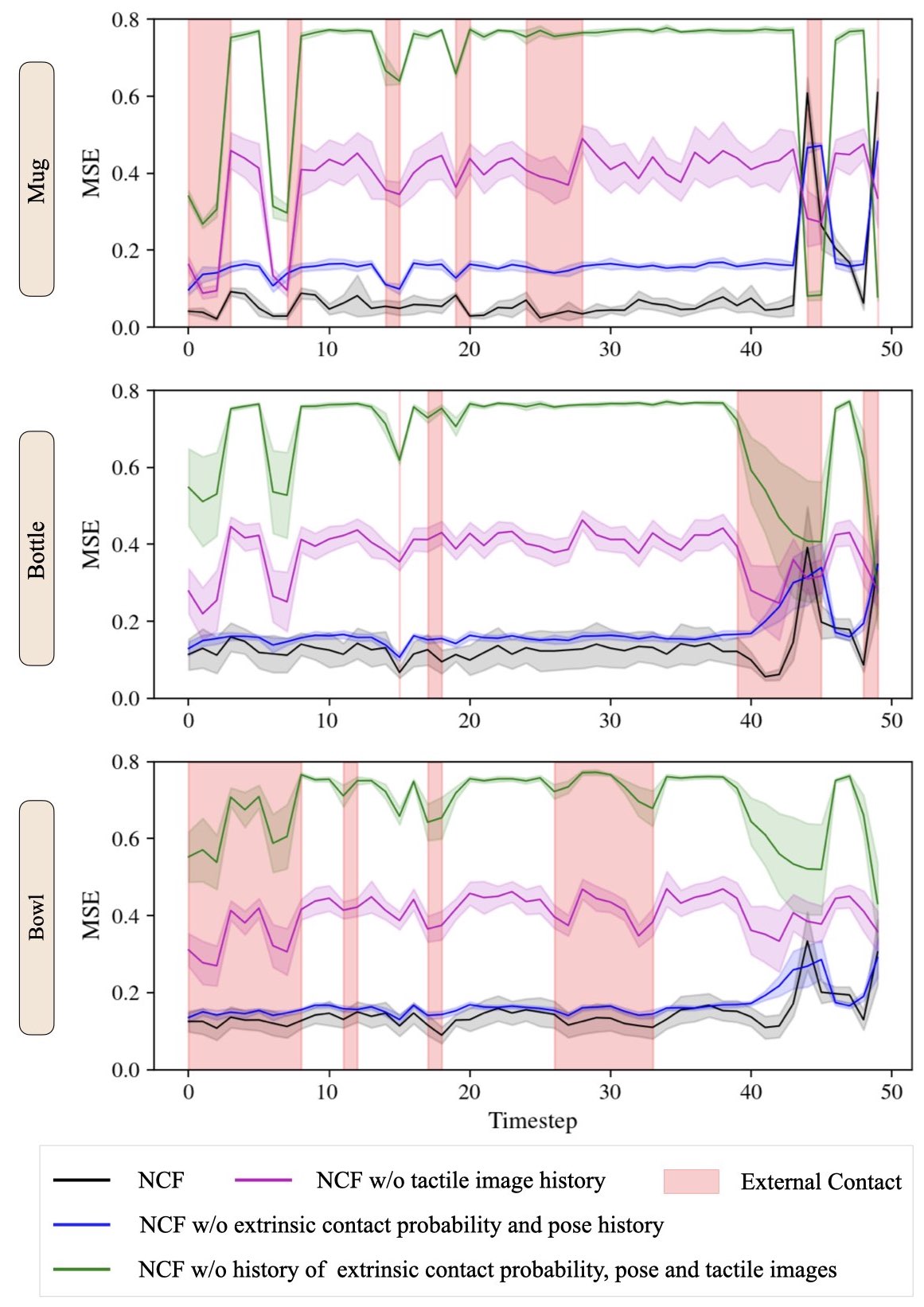}
    \vspace{-4mm}
    \caption{Ablations over inputs for tracking extrinsic contact with \ncf. We compare MSE with respect to ground-truth external contact probabilities over representative trajectories. We evaluate four models: \ncf, without extrinsic contact probability and end-effector pose history, without tactile images history (considering only the current frame), and without history of tactile images, extrinsic contact probability and end-effector pose. From previous timesteps considering extrinsic contact and sequences of tactile measurements significantly improves performance of the model.}
    \label{fig:ablations}
    \vspace{-6mm}
\end{figure}

In addition, we perform ablations over the inputs of our \ncf to quantify their contribution on the performance of the model. We compare four models: 1) \ncf; 2) \ncf without a history of contact probabilities and end-effector poses; 3) \ncf  considering only the current tactile frame rather than a history of images; and 4) \ncf without a history of tactile images, contact probabilities and end-effector poses. In~\cref{fig:ablations} we plot the MSE over time for each model and we highlight the timesteps with external contact events. The contribution of a history of tactile images for tracking extrinsic contact is noticeable. This is expected, given the intuition that the sequences of tactile images contain indirect information about the external contact. This results also suggest that considering only a history of contact probabilities instead of the sequence of tactile images generates a loss of information about how the external contact has been evolving over time. 

% seems like the sequence of tactile images helps to supply the history of contact probabilities. I think this is also expected, given that the  motion experienced by the object product of the contact is capture in those images and typically the external contact changes between consecutive timsteps are smooth. This is not completely the case for the other way around. Having a history of contact probabilities instead of the sequence of tactile images implies a loss of information about how the results of the object interacting with the environment.

\begin{figure}
    \centering
    \includegraphics[scale=0.315]{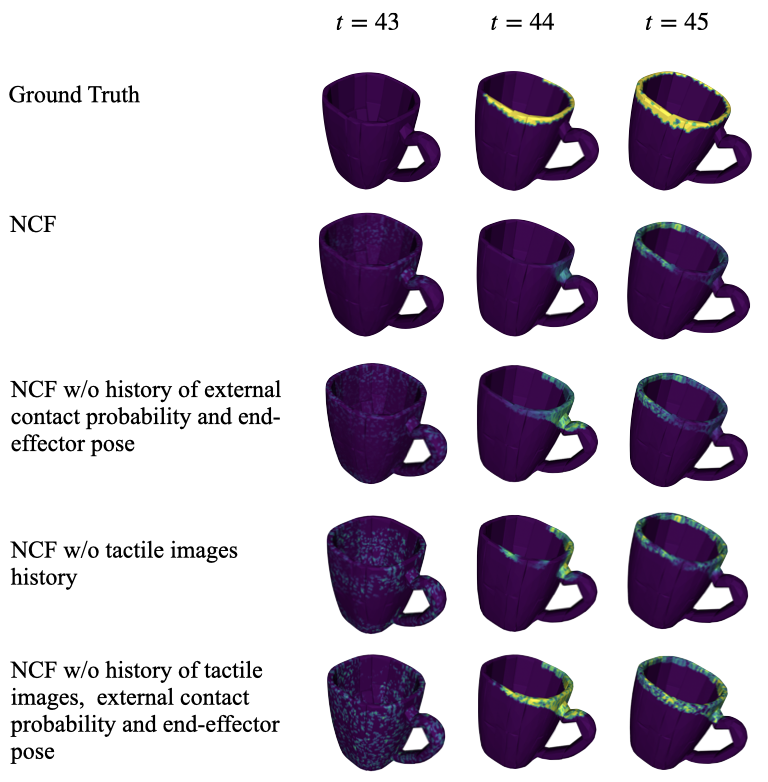}
    \vspace{-6mm}
    \caption{Illustration of failure case tracking extrinsic contact in the mug's test trajectory. We show the context around time step 44, where \ncf presents the highest error. In general, when transitioning from a no-contact state all models have a high error while the contact information starts propagating.}
    \label{fig:failure_case}
    \vspace{-7mm}
\end{figure}

We illustrate in~\cref{fig:failure_case} a failure case, specifically what is happening around the timestep where \ncf presents the highest error for the mug's test trajectory. We plot the extrinsic contact location prediction made by each of the four variants in the ablation and the ground truth for reference. In general, when transitioning from a no-contact state (in the figure, from timestep 43 to 44), all models have a high error while the contact information starts propagating. Note for example that the prediction made by the \ncf model at timestep 44 does not capture the contact interaction yet, but starts being noticeable at the next timestep. Additionally, by looking at the predictions of all models at timestep 43, we can notice that the models without tactile image history suffer from higher uncertainty when the object is not making contact. This suggest the advantage of using a sequence of tactile images instead of only the current frame.

%%%%%%%%%%%%%%%%%%%%%%%%%%%%%%%%%%%%%%%%%%%%%%%%%%%%%%%%%%%%%%%%%%%%%%%%%%%%%%%
\section{Discussion}
\vspace{-1mm}

\textbf{Summary.} We presented Neural Contact Fields (\ncf), a novel method for tracking extrinsic contact. Our model outputs the probability of external contact for a set of query points on the 3D surface of an object based on vision-based tactile sensing. Our method works in the latent space of spatial descriptors, which allow us to generalize within categories of objects with different shapes. Our experiments in simulation demonstrate the capability of \ncf in localizing and tracking complex contact interactions, such as multiple contact patches, and breaking and making of contact. We believe \ncf can be used in downstream robot manipulation tasks and leave its applications for future work.

\textbf{Limitations.} \ncf has been tested in simulation only and we anticipate sim2real will be nontrivial since simulation isn't perfect and acquiring ground-truth extrinsic contact labels on real data at scale will be intractable. To close the gap in simulation we can leverage for example, domain randomization over background and lighting of the DIGIT sensor. Another alternative can be fine-tuning the sim-trained \ncf on the real system with a downstream task policy in the loop. The current implementation of \ncf is limited to tracking extrinsic contacts for three classes of objects that are grasped with a fixed relative pose and a rigid grasp (no slipping). Given initial evidence it should be possible to scale training to larger datasets with diverse objects, which will additionally aid in addressing challenges to transferring to the real world. However, for relaxing the rigid grasp assumption, we will need to estimate the relative hand-object pose. This will allow us to study the ambiguity on where the contact might happen based on the grasp pose. Our model also requires known class of object in order to input the correct reference point cloud. Although this can be addressed using pre-trained object detection models, the correctness of \ncf will be compromised if the object is miss-classified. Finally, we are currently using the ground truths for establishing the prior information about the contact at the first timestep. This can be relaxed if during training we assume no-contact as prior for the first timestep of the trajectories.

%%%%%%%%%%%%%%%%%%%%%%%%%%%%%%%%%%%%%%%%%%%%%%%%%%%%%%%%%%%%%%%%%%%%%%%%%%%%%%%
% \addtolength{\textheight}{-12cm}   % This command serves to balance the column lengths
                                  % on the last page of the document manually. It shortens
                                  % the textheight of the last page by a suitable amount.
                                  % This command does not take effect until the next page
                                  % so it should come on the page before the last. Make
                                  % sure that you do not shorten the textheight too much.

%%%%%%%%%%%%%%%%%%%%%%%%%%%%%%%%%%%%%%%%%%%%%%%%%%%%%%%%%%%%%%%%%%%%%%%%%%%%%%%%

%%%%%%%%%%%%%%%%%%%%%%%%%%%%%%%%%%%%%%%%%%%%%%%%%%%%%%%%%%%%%%%%%%%%%%%%%%%%%%%%

%%%%%%%%%%%%%%%%%%%%%%%%%%%%%%%%%%%%%%%%%%%%%%%%%%%%%%%%%%%%%%%%%%%%%%%%%%%%%%%%

\section*{Acknowledgment}
The authors thank Sudharshan Suresh for feedback on the paper draft.

%%%%%%%%%%%%%%%%%%%%%%%%%%%%%%%%%%%%%%%%%%%%%%%%%%%%%%%%%%%%%%%%%%%%%%%%%%%%%%%%

\balance
\bibliographystyle{IEEEtran}
\bibliography{IEEEabrv, references}

\end{document}